# L'ORTHOGLIDE : UNE MACHINE-OUTIL RAPIDE D'ARCHITECTURE PARALLELE ISOTROPE

**Philippe Wenger**

Institut de Recherche en Communications et Cybernétique de Nantes UMR CNRS 6597, 1 rue la Noë BP 92101 44321 Nantes cedex – Tel. +33 (0)240376947, Fax +33 (0)240376930, Philippe.Wenger@irccyn.ec-nantes.fr

**Damien Chablat**

Institut de Recherche en Communications et Cybernétique de Nantes UMR CNRS 6597, 1 rue la Noë BP 92101 44321 Nantes cedex – Tel. +33 (0)240376947, Fax +33 (0)240376954, Damien.Chablat@irccyn.ec-nantes.fr

**Félix Majou**

Département de Génie Mécanique, Université Laval, Université Laval, Québec, Québec, Canada, G1K 7P4, felix@gmc.ulaval.ca (jusqu'au 30/08/2002)

Institut de Recherche en Communications et Cybernétique de Nantes UMR CNRS 6597, 1 rue la Noë BP 92101 44321 Nantes cedex – Tel. +33 (0)240376947, Fax +33 (0)240376930, Felix.Majou@irccyn.ec-nantes.fr (à partir du 1/09/2002)

**Résumé :**

Cet article présente le projet "Orthoglide" de l'IRCCyN. Ce projet a pour but la réalisation d'un prototype de machine-outil rapide à trois degrés de translation. La particularité de cette machine est une architecture cinématique parallèle optimisée pour obtenir une volume de travail compact et où les performances sont homogènes. Pour cela, le critère principal de conception qui a été utilisé est l'isotropie. Le prototype expérimental destiné principalement à valider la cinématique est en cours d'assemblage. Il possède un volume de travail cubique de 200mm de coté, pour une emprise au sol de 1m². Par rapport aux machines UGV industrielles, son échelle est réduite de façon a pouvoir la déplacer facilement. Ce prototype permettra d'usiner des petites pièces plastique (type prototypage rapide). La vitesse et l'accélération maximales de l'outil sont respectivement de 1m/s et 20m/s². Ce projet a reçu le soutien de l'ANVAR et de la région Pays-de-Loire. La commande et l'étalonnage sont en cours d'étude dans le cadre d'un projet interdisciplinaire de recherche du CNRS. Une version 5 axes de l'orthoglide est également à l'étude. L'équilibrage statique et dynamique sera abordé dans le cadre d'une collaboration avec l'Université Laval de Québec.

**Mots clés : Conception, Machine parallèle, isotropie, prototype, UGV**

## 1 Introduction

L'usinage à grande vitesse (UGV) exige des performances dynamiques de plus en plus élevées de la part des machines-outils. Ces performances peuvent être améliorées en équipant les machine-outil de moteurs plus puissants. Cependant, ces améliorations sont limitées par les masses élevées des axes des machines-outils classiques dites « sérielles » (axes en séries). Sur une machine-outil sérielle, les axes sont « empilés » les uns sur les autres. Sur la figure 1 par exemple, l'axe Y supporte l'axe X. Le moteur de l'axe Y doit donc déplacer deux corps massifs.

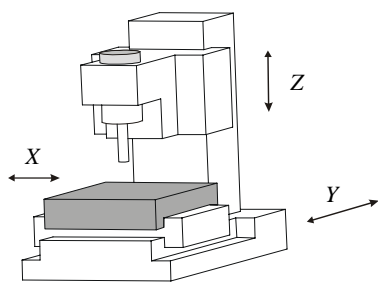

Fig. 1 : *Machine-outil sérielle*

Pour diminuer les inerties des machines-outils, une solution très intéressante consiste à changer l'architecture cinématique en plaçant les axes non pas en série mais en parallèle (architecture parallèle) [1]. La première application industrielle connue des mécanismes parallèles est la plate forme de Gough [2] (Figure 2), destinée au test de pneumatiques. C'est un mécanisme à six degrés de liberté, muni de six vérins reliant la base à la plate-forme mobile. A la fin des années soixante, D.Stewart réutilisera cette architecture pour concevoir un simulateur de vol [3], [4].

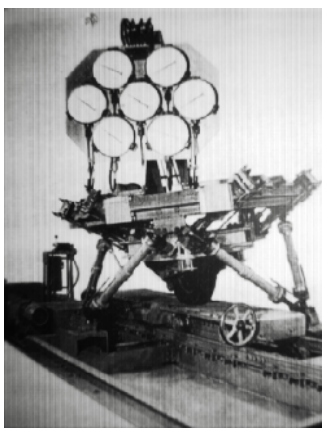

Fig. 2 : *Plate-forme de Gough-Stewart*

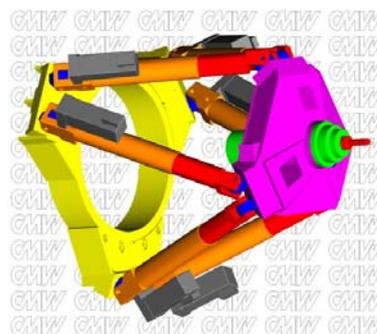

Fig. 3 : *Machine-outil parallèle « hexapode » (doc. CMW)*

La première présentation officielle d'une machine-outil parallèle remonte à 1994 avec la Variax (Gidding & Lewis) au salon de Chicago. Cette machine reprenait l'architecture de type plate-forme de Gough-Stewart communément appelée « hexapode ». Depuis cette date, plusieurs industriels travaillent sur des prototypes de machines-outils parallèles. La plupart d'entre eux ont repris l'architecture hexapode. La figure 3 montre un exemple de machine-outil parallèle hexapode de la société CMW. L'outil est relié à une base fixe au moyen de six jambes télescopiques montées en parallèle. Les masses en mouvement sont plus faibles que dans une machine-outil série puisque chaque moteur ne déplace que le plateau supportant l'outil. De plus, les jambes ne subissant aucune contrainte de flexion, leur structure peut être allégée. En revanche, les limites des architecture hexapode, comme de la plupart des architectures parallèles, sont un volume de travail restreint et complexe. Les équations qui relient les déplacements de l'outil à ceux des moteurs sont non linéaires ce qui engendre des variations importantes des performances au sein du volume de travail.

Une alternative à l'architecture hexapode a été présentée par l'ETH Zürich avec l'hexaglide. Cette architecture se caractérise par des jambes de longueur fixe qui glissent sur des rails (Figure 4). L'avantage de cette architecture réside dans le fait que les moteurs sont fixes, ce qui diminue les inerties et permet l'emploi de moteurs linéaires. De plus, la dissipation thermique des moteurs est facilitée.

Des versions 3 axes de l'hexaglide ont été proposées par la suite avec le Triaglide (Mikron), le Linapod (ISW Uni Stuttgart), le Quickstep (Krause & Mauser) et l'Urane SX (Renault-Automation). Ces machines reprennent en fait une architecture déjà proposée en robotique avec par exemple le « Delta linéaire » [5] et le « Y-Star » [6] permettant de maintenir une orientation fixe de l'outil (figure 5).

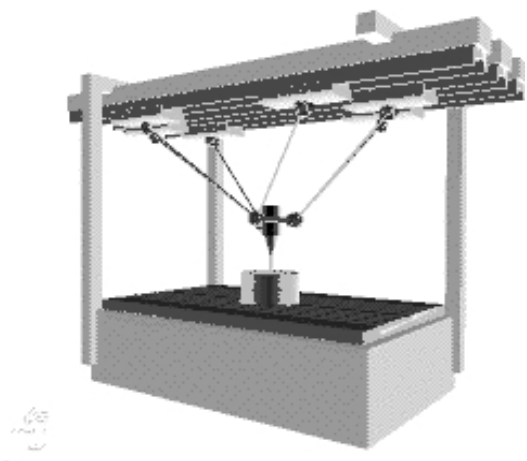

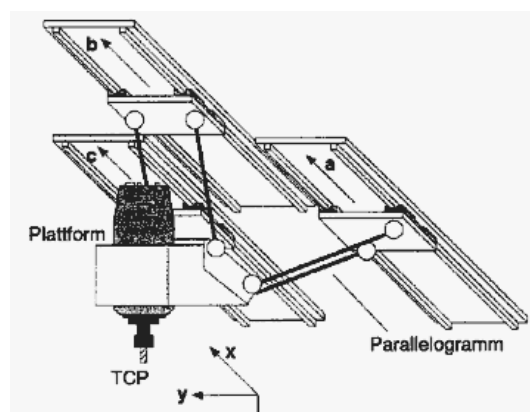

Fig. 4 : *Machine "Hexaglide"*          Fig. 5 *: Machine "Triaglide"*

Les machines hexaglide et triaglide ont pour avantage supplémentaire d'offrir un volume de travail présentant une dimension ajustable à la demande (il suffit d'augmenter la longueur des liaisons glissières pour augmenter d'autant la longueur du volume de travail). Cependant, les performances restent non homogènes dans le volume de travail. Par exemple, la rigidité dans la direction verticale est différente de celle dans la direction horizontale et varie fortement en fonction de la position en z de l'outil [7].

Il existe d'autres structures de machines parallèles, comme par exemple des versions hybrides parallèle-sérielle à l'image de la machine 5 axes Tricept 805 du suédois Neos-Robotics, ou du prototype 3 axes de l'IFMA [8]. Nous ne les citerons pas toutes dans cet article. Pour une vision très complète des machines parallèles existantes, le lecteur pourra se reporter au site internet [9] particulièrement bien documenté sur ce sujet.

## 2. Description de l'orthoglide

### 2.1 Cahier des charges

Le cahier des charges qui a guidé l'élaboration de l'orthoglide est le suivant. L'objectif était la conception d'une machine 3 axes rapide d'architecture parallèle, extensible à 5 axes, ne présentant pas les inconvénients inhérents aux mécanismes parallèles. Les critères principaux de conception qui ont été retenus sont les suivants :

- 3 actionneurs fixes de type glissières (diminution des inerties, possibilité d'utiliser des moteurs linéaires, meilleure dissipation thermique)
- volume de travail de forme régulière proche d'un cube
- homogénéité des performances dans tout le volume de travail et dans toutes les directions
- symétrie de construction (diminution des coûts)
- articulations simples (pas de cardan ni de rotule)

Un prototype peu onéreux à échelle réduite devait être réalisé en fonction de ce cahier des charges pour valider la cinématique et notre méthode de conception.

### 2.2 Choix de la cinématique de base

Compte tenu des objectifs cités précédemment, la cinématique de base repose sur un ensemble de trois liaisons glissières motorisées qui actionnent l'outil selon un mouvement de translation spatiale. L'outil est relié aux actionneurs par le moyen de trois jambes identiques constituées de parallélogrammes articulés (figure 6). La figure 7 montre l'architecture cinématique de l'orthoglide, représentée dans la configuration isotrope (voir §3).

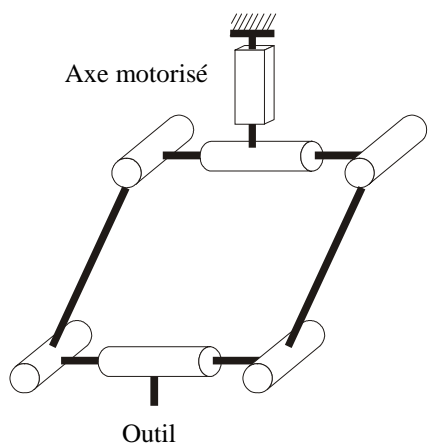

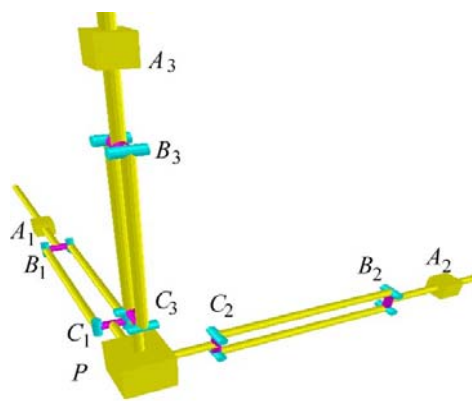

Fig. 6 : *Cinématique d'une jambe*          Fig. 7 : *Cinématique de l'orthoglide*

Le choix de l'agencement orthogonal des liaisons glissières motorisées est justifié par le critère d'isotropie et d'homogénéité des performances (voir §3.1 suivant). L'étude cinématique détaillée de l'orthoglide est décrite dans [10].

## 3. Principe de conception et d'optimisation géométrique

### 3.1. Manipulabilité, conditionnement et isotropie

Pour traduire l'homogénéité des performances dans le volume de travail, il existe un indice particulièrement bien adapté qui s'appuie sur la matrice jacobienne du mécanisme parallèle : la *manipulabilité* [11]. Rappelons que la matrice jacobienne d'un mécanisme relie les vitesses du solide à contrôler à celle des actionneurs. Pour les mécanismes parallèles, on utilise plutôt la jacobienne inverse $\mathbf{J}^{-1}$, d'écriture plus simple. On a alors la relation suivante :

$$\dot{\boldsymbol{\rho}} = \mathbf{J}^{-1} \, \dot{\mathbf{p}} \tag{1}$$

où $\dot{\boldsymbol{\rho}}$ est le vecteur des vitesses des moteurs et $\dot{\mathbf{p}}$ est le vecteur des vitesses de l'outil.

La manipulabilité peut être définie en vitesses ou en efforts, les deux notions étant duales. La manipulabilité peut être représentée par un ellipsoïde. Celle-ci traduit les distorsions d'une sphère unité sous l'effet du transformateur de coordonnées entre les vitesses des moteurs et celles de l'outil (ou entre les efforts exercés par les moteurs et les efforts exercés par l'outil). Cette distorsion fait que la distribution des vitesses (ou des efforts) de l'outil autour d'un point donné n'est pas uniforme. On peut mesurer cette distorsion par le rapport entre la longueur du petit axe et celle du grand axe de l'ellipsoïde de manipulabilité. Ce rapport est directement lié au *conditionnement* de la matrice jacobienne qui s'écrit comme le rapport entre la plus petite et la plus grande valeur singulière de la matrice jacobienne [12]. Physiquement, ces valeurs singulières représentent les facteurs de transmission des vitesses (ou des efforts) dans les directions des axes principaux de l'ellipsoïde de manipulabilité. Dans le cas d'une machine sérielle, les relations d'entrée-sortie sont linéaires : il n'existe pas distorsion. De plus, les facteurs de transmission des vitesses valent 1 : une vitesse de 1m/s d'un moteur se traduit pas une vitesse identique de l'outil dans la direction de ce moteur. Dans tout le volume de travail, l'ellipse de manipulabilité est alors une sphère unité. Ceci n'est pas le cas des machines parallèles car les relations d'entrée-sortie sont non linéaires. Il est toutefois possible de limiter les distorsions en imposant des limites sur le conditionnement de la matrice jacobienne, et en imposant l'existence d'une configuration isotrope de la machine.

### 3.2. Existence d'une configuration isotrope

La condition d'existence d'une configuration isotrope dans le volume de travail de l'orthoglide s'obtient en écrivant que la matrice jacobienne inverse $\mathbf{J}^{-1}$ est proportionnelle à la matrice identité. Cette matrice s'écrit [10] :

$$\mathbf{J}^{-1} = \begin{bmatrix} \dfrac{1}{\eta_1}\,(\mathbf{c}_1 \text{ - } \mathbf{b}_1)^T \\[2mm] \dfrac{1}{\eta_2}\,(\mathbf{c}_2 \text{ - } \mathbf{b}_2)^T \\[2mm] \dfrac{1}{\eta_3}\,(\mathbf{c}_3 \text{ - } \mathbf{b}_3)^T \end{bmatrix} \text{ avec } \eta_i = (\mathbf{c}_i\text{-}\mathbf{b}_i)^T \dfrac{\mathbf{b}_i\text{-}\mathbf{a}_i}{\|\mathbf{b}_i\text{-}\mathbf{a}_i\|} \tag{2}$$

Les vecteurs $\mathbf{a}_i$, $\mathbf{b}_i$ et $\mathbf{c}_i$ représentent la position des points $A_i$, $B_i$ et $C_i$ définis en figure 7.

Pour que la matrice $\mathbf{J}^{-1}$ soit proportionnelle à la matrice identité, il suffit que :

$$\frac{1}{\eta_1}\,\|\mathbf{c}_1 \text{ - } \mathbf{b}_1\| = \frac{1}{\eta_2}\,\|\mathbf{c}_2 \text{ - } \mathbf{b}_2\| = \frac{1}{\eta_3}\,\|\mathbf{c}_3 \text{ - } \mathbf{b}_3\| \tag{3}$$

et $\qquad (\mathbf{c}_1 \text{ - } \mathbf{b}_1)^T\,(\mathbf{c}_2 \text{ - } \mathbf{b}_2) = 0,\ (\mathbf{c}_2 \text{ - } \mathbf{b}_2)^T\,(\mathbf{c}_3 \text{ - } \mathbf{b}_3) = 0,\ (\mathbf{c}_3 \text{ - } \mathbf{b}_3)^T\,(\mathbf{c}_1 \text{ - } \mathbf{b}_1) = 0 \qquad (4)$

La première équation traduit le fait les angles entre les glissières et les parallélogrammes sont les mêmes pour chaque jambe. La seconde équation traduit que les parallélogrammes sont orthogonaux entre eux. A ce stade, on ne peut pas encore en déduire de condition sur l'orientation des glissières. Il faut pour cela écrire que dans la configuration isotrope, la matrice jacobienne s'écrit exactement comme la matrice identité (comme pour une machine sérielle).

### 3.3. Condition sur les facteurs de transmission à l'isotropie

Pour que dans la configuration isotrope, l'orthoglide ait le même comportement qu'une machine sérielle, on doit donc écrire que dans cette configuration, les facteurs de transmission des vitesses valent 1. Pour cela, il faut que les trois termes de l'équation (3) valent 1 :

$$\frac{1}{\eta_1}\,\|\mathbf{c}_1 \text{ - } \mathbf{b}_1\| = \frac{1}{\eta_2}\,\|\mathbf{c}_2 \text{ - } \mathbf{b}_2\| = \frac{1}{\eta_3}\,\|\mathbf{c}_3 \text{ - } \mathbf{b}_3\| = 1 \tag{5}$$

Ceci signifie que pour chaque jambe, l'axe du parallélogramme doit être aligné avec la liaison glissière. Ajoutée à la condition d'orthogonalité des parallélogrammes dans la configuration isotrope, la condition (5) implique que les glissières doivent être orthogonales (d'où le nom de « orthoglide »).

### 3.4. Optimisation géométrique en fonction d'un volume de travail prescrit

La dernière étape de notre démarche de conception est l'optimisation des dimensions de l'orthoglide en fonction d'un volume de travail prescrit (longueurs des jambes, position et débattement des liaisons glissières).

Conformément au cahier des charges, nous décrivons le volume de travail prescrit comme un cube de dimension donnée $L_w$, au sein duquel les facteurs de transmissions doivent être bornés par des limites données, de façon à réduire les dispersions des performances (figure 8).

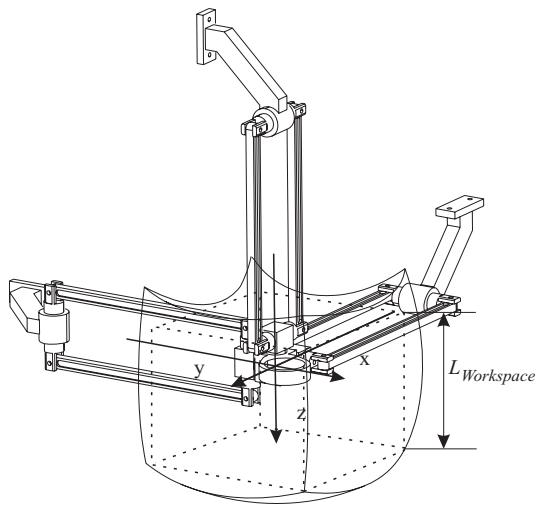
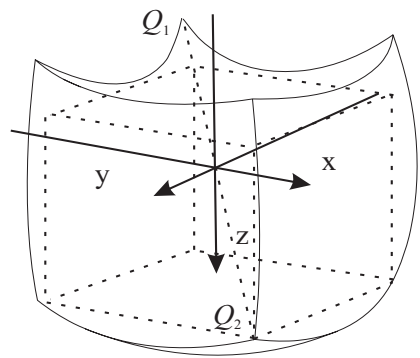

Fig. 8 : *Volume de travail prescrit*          Fig. 9 : *Définitions des points de référence*

Pour cela, nous mettons en évidence deux points de référence $Q_1$ et $Q_2$ (Figure 9) tels que si ces points peuvent être atteints par l'outil avec des facteurs de transmissions bornés par les limites imposées, l'ensemble du cube défini à partir de ces points est accessible et respecte ces limites. Ceci permet de réduire l'étude à ces deux points. La justification scientifique de ce résultat est exposée dans [13]. Ces points de références sont alors utilisés pour définir les longueurs des jambes (conformément au cahier des charges, nous imposons des longueurs de jambes identiques pour des raisons de symétrie) en fonction du coté $L_w$ du cube prescrit. Puis, les positions et les débattement des glissières sont calculés en écrivant les équations du modèle géométrique direct aux deux points de référence. Les détails de ces calculs sont présentés dans [13].

## 4. Prototype

Un prototype à l'échelle réduite est en cours de réalisation. Le volume de travail prescrit est un cube de coté $L_w$=200mm. Les limites inférieures et supérieures des facteurs de transmission des vitesse que nous avons imposées sont respectivement de 0.5 et 2. Les dimensions que nous avons obtenues sont de 310 mm pour la longueur des jambes et de 257 mm pour le débattement des liaisons glissières. Le rapport entre la taille du cube prescrit et les débattements articulaires est de r=200/257=0.78. Ce résultat confirme la bonne compacité offerte par l'architecture orthogonale des glissières, qui avait déjà été observée dans une version 2 axes de l'orthoglide [14], [15]. Notons que ce rapport augmente lorsque l'on éloigne les limites sur les facteurs de transmission des vitesses (par exemple en fixant 1/3 et 3 au lieu de 0.5 et 2). Les actionneurs sont réalisés par des vis à billes entraînées par des moteurs rotatifs (des moteurs linéaires pourront être employés par la suite). Les moteurs ont été calculés pour fournir une vitesse et une accélération de 1.2 m/s et 20 m/s$^2$ dans la configuration isotrope. La figure 10 suivante montre le modèle CAO du prototype, la figure 11 montre un gros plan sur le prototype assemblé.

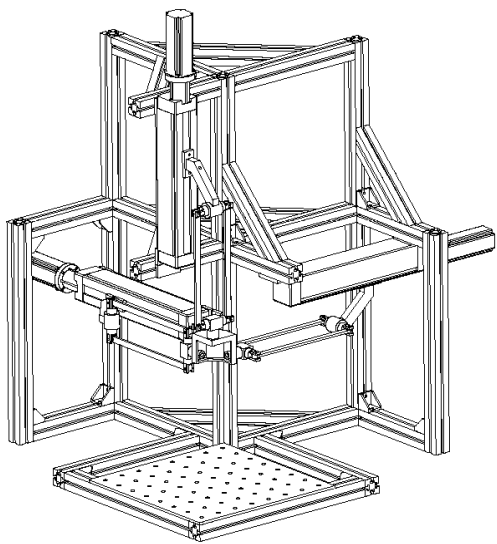 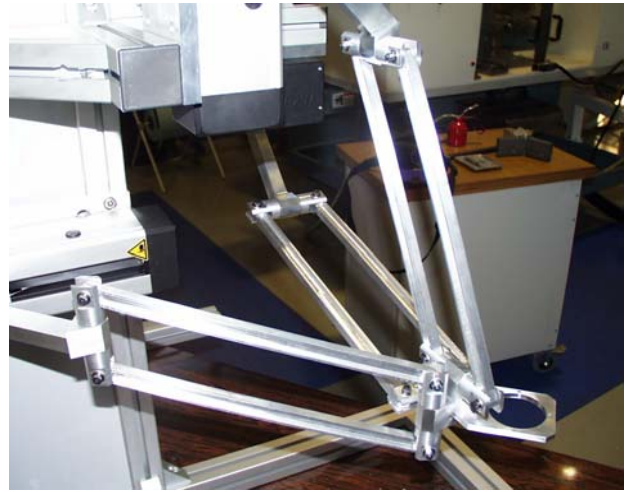

Fig. 10 : Modèle CAO du prototype          Fig. 11 : Réalisation du prototype

## 5. Conclusion

Le but du projet orthoglide était de valider une nouvelle cinématique parallèle de machine-outil rapide. Cette cinématique se caractérise par une bonne compacité, un volume de travail quasi-cubique et des variations limitées des performances au sein de ce volume de travail. La machine possède 3 jambes identiques et toutes les articulations sont simples. Une méthodologie de conception a été élaborée de façon à pouvoir calculer les dimensions de la machine en fonction de la taille d'un cube à atteindre et de limites imposées sur les facteurs de transmission des vitesses. Les perspectives à court terme de ce projet sont la mise en œuvre de la commande et l'étude du comportement dynamique du prototype. Un modèle dynamique symbolique a déjà été élaboré. La commande et l'étalonnage sont en cours d'étude dans le cadre d'un projet interdisciplinaire de recherche du CNRS. Dans un second temps, une optimisation structurelle sera conduite en confrontant des résultats expérimentaux avec des résultats de simulations en cours sur le logiciel Mecano que nous menons en partenariat avec l'Université de Liège. D'autre part, l'équilibrage statique et dynamique sera abordé dans le cadre d'une collaboration avec l'Université Laval de Québec.

A plus long terme, une version pré-industrielle du prototype est envisagée avec l'aide de l'ANVAR et une version 5 axes sera proposée.

## Références